\documentclass[letterpaper, 10 pt, conference]{ieeeconf}  
\IEEEoverridecommandlockouts                              
\usepackage{graphics}    
\usepackage{times}       
\usepackage{amsmath}     
\usepackage{amssymb}     
\usepackage{graphicx}
\usepackage{algorithm}
\usepackage[noend]{algpseudocode}

\usepackage{siunitx}
\usepackage{booktabs}
\usepackage{comment}
\usepackage{float}

\def\eqref#1{Eq.~(\ref{#1})}

\newcommand\todo[1]{\textbf{[TODO: #1}]}
\newcommand\etal{\emph{et al.}}

\usepackage[font=small]{caption}

\title{\LARGE \bf Sound Matters: Auditory Detectability of Mobile Robots}

\author{Subham Agrawal* \and Marlene Wessels* \and Jorge de Heuvel \and Johannes Kraus \and Maren Bennewitz
    \thanks{* These authors contributed equally to this work.}
    \thanks{Subham Agrawal, Jorge de Heuvel, and Maren Bennewitz are with the Humanoid Robots Lab at the University of Bonn,
Germany. Jorge de Heuvel and Maren Bennewitz are also  with the Lamarr Institute for Machine Learning and Artificial Intelligence, Bonn, Germany.}%
    \thanks{Marlene Wessels and Johannes Kraus are with the Institute of Psychology, Experimental Psychology, Johannes Gutenberg-University Mainz, Germany.}%
    \thanks{This work has partially been funded by the Deutsche Forschungsgemeinschaft (DFG, German Research Foundation) under the grant number \mbox{BE~4420/2-2}~(FOR 2535 Anticipating Human Behavior).}
}

\begin{document}
\maketitle
\thispagestyle{empty} 
\pagestyle{empty}

\begin{abstract} 
Mobile robots are increasingly being used in noisy environments for social purposes, e.g. to provide support in healthcare or public spaces. 
Since these robots also operate beyond human sight, the question arises as to how different robot types, ambient noise or cognitive engagement impacts the detection of the robots by their sound. 
To address this research gap, we conducted a user study measuring auditory detection distances for a wheeled (Turtlebot 2i) and quadruped robot (Unitree Go 1), which emit different consequential sounds when moving. Additionally, we also manipulated background noise levels and participants' engagement in a secondary task during the study. 
Our results showed that the quadruped robot sound was detected significantly better (i.e., at a larger distance) than the wheeled one, which demonstrates that the movement mechanism has a meaningful impact on the auditory detectability. The  detectability for both robots diminished significantly as background noise increased. But even in high background noise, participants detected the quadruped robot at a significantly larger distance. The engagement in a secondary task had hardly any impact. 
In essence, these findings highlight the critical role of distinguishing auditory characteristics of different robots to improve the smooth human-centered navigation of mobile robots in noisy environments.


\end{abstract} 

\section{Introduction}
\label{sec:intro}
\begin{figure}[t] 	
\centering 	
\includegraphics[width=0.8\linewidth]{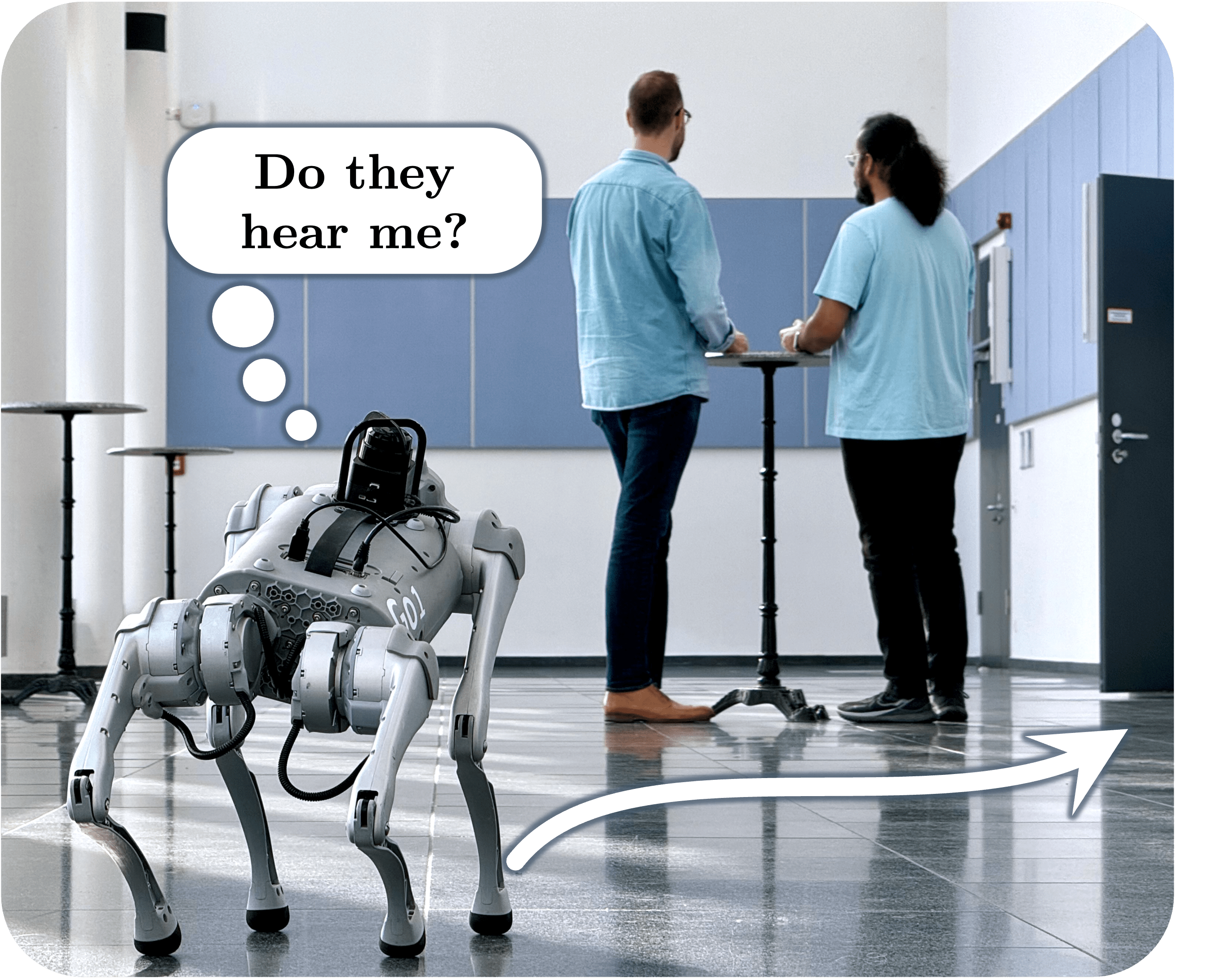} 	
\caption{ 		
        This study explores the auditory detectability of two mobile robots in low vs. high background noise, and with vs. without cognitive engagement of the human observers. Our findings serve as a valuable foundation for the intelligent design of HRI tailored to specific human needs.
		\label{fig:overviewImage}
	} 
 \end{figure}

The rapid advances in robotics and artificial intelligence enable an increasing integration of mobile robots into daily life, which increases the importance of human-robot interaction (HRI). 
The distance a robot maintains from humans is relevant for a successful and pleasant interaction \cite{kim2014social,kluber_keep_2023,de_heuvel_learning_2022}, but depends strongly on the interaction context.
In cooperative settings, humans need to engage closely with robots, such as when robots assist in household chores or act as companions. 
At other times such as during the task conduction of robots in public spaces (e.g., cleaning or deliveries \cite{babel2022findings}), humans take the role of merely co-present persons whose tasks are not directly intertwined with the one of the robots.
In these situations, the humans’ timely awareness of an approaching robot from all directions is necessary to prevent unexpected encounters and collisions.
Conversely, there are instances where humans prefer robots to execute tasks discreetly (Fig.~\ref{fig:overviewImage}), as seen in hospitals where patient-monitoring robots should operate inconspicuously to avoid disturbances \cite{han2017understanding}.
The sound emitted by the robot plays a meaningful role in this context. 
These sounds inherently produced by robots during operation are referred to as consequential sounds \cite{van2008experience}. 
Since the sound intensity amplifies as the robot approaches the human, this allows for a natural and intuitive interface signaling an approaching robot.

However, the sound profiles vary among different types of robots due to different movement mechanisms. 
For instance, the wheeled Turtlebot 2i produces a continuous high-frequency consequential sound, while the quadruped Unitree Go 1 generates a rhythmic alternation of lower frequency components resembling an impact sound. 
Despite having the same objective sound intensity, these distinct sound profiles may impact the subjective human auditory perception differently, so that an approach or the distance of certain robot types could be differently recognized.
Especially, the context (e.g., different background noises) could play a role which movement sounds are better suited in terms of providing optimal detectability.
Yet, in environments with high background noise, the differences in robot sound may become less significant. 
Additionally, when individuals are engaged in other tasks, they may be less attentive to the sound of the robot in general, potentially reducing the detectability of robots.
In sum, the type of robot, background noise levels, and engagement in secondary tasks emerge as factors potentially influencing the auditory detectability of robots. 
Understanding these factors is crucial for designing human-centered and needs-oriented navigation algorithms of mobile robots. 
Against this background, we conducted a controlled experiment inspired by classic approaches from perceptual psychology to investigate the impact of robot type, background noise and a secondary task on its detectability.

Our study addressed the following \textbf{three research questions}:
\noindent\textbf{1)} Does the auditory detectability vary between robots (wheeled vs. quadruped), which emit distinctively different consequential sounds? 
\noindent\textbf{2)} Does the auditory detectability for both robots suffer differently in noisy environments? 
\noindent\textbf{3)} Does the auditory detectability decrease when the user is engaged in another activity?




\section{Related Work}
\label{sec:related}

Previous studies investigated the influence of consequential sounds on HRI and showed negative effects on the interaction with robots due to these sounds, e.g., humans kept a greater distance to the robots \cite{robinson2020implicit, bhagya2019exploratory}. 
This effect can be mitigated by masking background sounds \cite{trovato2018sound} \cite{latupeirissa2023probing}.\\
Nonetheless, there are use cases in which prominent robot sounds are indispensable — particularly, when humans are not continuously interacting with a robot in the same task and might therefore lose track of its location or when humans are required to collaborate closely with robots and must therefore remain aware of the robot's proxemics, notably in healthcare and industrial settings. 
In fact, humans allowed a closer distance to robots when they could not only see but also hear the robot \cite{petrak2019let}.
Cha \etal~\cite{cha2018effects} found that while both broadband and tonal sounds improve auditory detectability and localization of robots, humans prefer broadband sounds as they are perceived as less irritating.
Regarding human-robot proxemics, Samarakoon \etal~\cite{samarakoon2022review} noted that a familiar environment complemented with audio leads to closer interaction, which highlights the human adaptability in HRI. 
Similarly, Johannsen \cite{johannsen2001auditory} demonstrated the effective use of natural robotic sounds blended with music for overseeing industrial processes in multimedia contexts.\\
The specific requirements for robot sound design are significantly influenced by the application context. 
Tsarouchi \etal~\cite{tsarouchi2016human} emphasized particularly the challenges of HRI in high-noise environments (up to 89 dB in some industrial settings). 
Such environments require sound design adaptations to ensure that robots remain detectable but at the same time do not add to unnecessary noise pollution, especially when robots operate beyond human sight.
Equipping robots with artificial sounds would be one possibility, similar as in the interaction of road users with electric vehicles \cite{moore2020sound}\cite{wessels_audiovisual_2022}.
A potential alternative approach could, however, be to consider the differences in consequential sounds emitted by different robot types. 
If the consequential sounds affect the human auditory detectability differently, it would be possible to define optimal use cases for each robot type and to consider these differences in robot navigation algorithms. 
That is, robots that can be detected by humans more easily (i.e. at lower sound intensities), even in the presence of high ambient noise, could be better suited for use cases with a higher distance between humans and robots like co-existence in public spaces or collaboration in industrial settings. 
Conversely, robots that can only be detected at very near distances, could be better suited for close HRI, where the robot should operate inconspicuously.\\

This suggests that understanding the detectability of consequential robot sounds, will facilitate the design of more human-centered, context-dependent HRI strategies without the strict necessity to implement complex sound systems for robots.
In this study, we therefore focused on these differences between consequential robot sounds in a scenario where the robot operates \textit{beyond} human sight. 
Based on our findings, we will provide practical guidance on which types of robots are better suited for specific contexts, thus contributing to the broader discussion on sound design in robotics and its impact on HRI.


\begin{figure*}[ht!] 	
\centering
\includegraphics[width=0.8\linewidth]{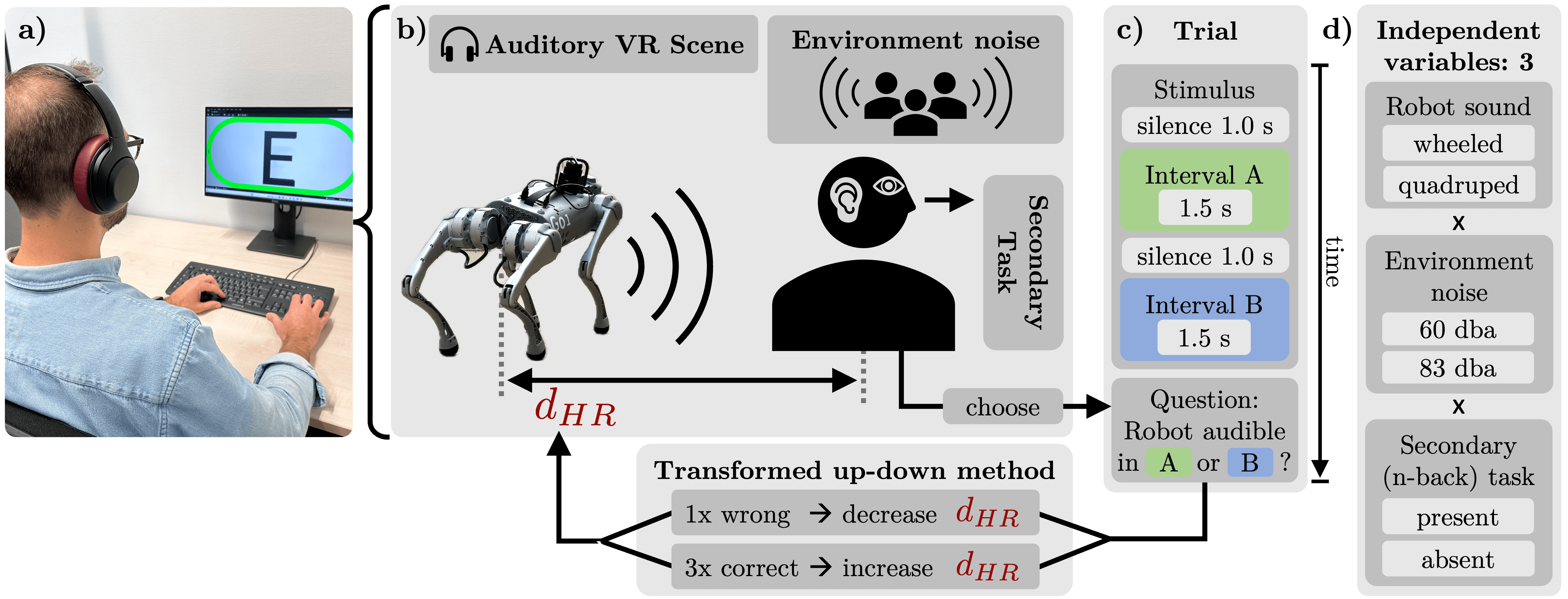}
\caption{Schematic of our user study design showing \textbf{a)} a photo of a participant conducting the experiment. \textbf{b)} Variables: We investigate the audible distance~$d_{HR}$ of robot detection beyond human sight based upon the independent variables IV1 consequential robot sound type, IV2 environment noise level and IV3 engagement of the participant in a secondary (one-back) task. The auditory detection distance of a robot for three independent variables is measured via the transformed up-down adaptive procedure. \textbf{c)} Characteristic of each trial: In one trial, the participant is first presented with an auditory stimulus followed by a question. Each auditory stimulus, consists of two intervals A and B of \SI{1.5} {\second} each, each preceded by a \SI{1.0} {\second} silence, respectively. Note that the robot sound is only presented in one of the intervals. The user has to tell apart the audibility of the robot from the environment noise by choosing the interval that presumably contained the robot sound. \textbf{d)} IV levels: All 8 combinations of the 3 independent variables are presented in an interleaved, random manner to the participant. They are, however, grouped and presented in two different blocks - with and without secondary task to make it simpler to answer workload related question for the participants.}
\label{fig:user_study_design}
\end{figure*} 


\section{Methodology}
\label{sec:methodology}

This section outlines the approach employed to investigate the detectability of robot sounds in two background noise environments, with and without participants being engaged in a secondary task.
The robot sounds were recorded using binaural microphones (files available on our website\footnote{https://www.hrl.uni-bonn.de/publications/2024/agrawal24roman/}) and reproduced in a virtual environment for two tasks: 1) a loudness matching task to adapt the intensity of the robot sounds, so that they were perceived at equal subjective loudness, and 2) a detection task, which determined at which distance humans detected each robot sound reliably.
An overview and breakdown of the detection task is presented in Fig.~\ref{fig:user_study_design}.

\subsection{Robot Sound Simulation}
The simulation of the robots sounds in the virtual environment was based on acoustic recordings of real robots. 
After calibrating the requisite hardware components, we used a 3DIO FS XLR Binaural Microphone to capture the consequential sounds of two robot types — namely, the Kobuki Turtulebot 2i (wheeled robot) and the Unitree Go 1 (quadruped robot).
The recording of the robot sound transpired at a height of 7 cm from the ground while the robot moved. 
The wheeled robot performed a back and forth maneuver at a mean distance of 1m and a speed of 0.1 m/s directed towards the position of the microphone. 
The quadruped robot's sound was recorded at a fixed distance of 1 m due to its ability of walking at the same location.
We determined the respective amplitude levels. 
Subsequently, leveraging the pertinent head-related transfer functions (HRTFs) \cite{geronazzo2018impact, cheng2001introduction}, we reproduced the robot sounds within the virtual context inside Unreal Engine \cite{unrealengine} using spatial audio rendering techniques through the Resonance Audio plugin from Google \cite{resonanceaudio}.
The acoustic signals were channeled through a pair of Sony WH-XB900N headphones and fed back into dummy ears with binaural microphones. 
We modulated the playback intensity, so that it matched the original recorded sound amplitudes. 
The calibration process ensured the creation of a consistent auditory environment for the subsequent phases of the investigation.\\
This simulation approach enabled us to adjust the sound intensity of each robot sound, so that it corresponded to a specific distance. 
Hence, it was possible to simulate the propagation of the robot sound from any distance without having to record the robot sound from each of these distances.
We adjusted the sound intensity linearly in the detection task, which is explained in detail in the following sections.
Note however that the sound intensity decreases logarithmically with the distance from the source, as described by the inverse square law of sound intensity in a free field \cite{hartmann_signals_1998}, stating that $I \propto 1/d^2$, where $I$ is the intensity of the sound wave at a distance of $d$ from the source.
Therefore, we determined the distances corresponding to the respective sound intensities in the detection task by applying a logarithmic transformation function according to the inverse square law of sound intensity.

\subsection{Background Noise Simulation}
Taking into account common operating environments for mobile robots, we chose a combination of a factory environment characterized by machine noise and a large crowd of people speaking indistinctly as the background noise
After mixing the background noise, we adjusted the output at two standardized sound intensity levels:
Based on the guidelines of the Center for Hearing and Communication\footnote{https://noiseawareness.org/info-center/common-noise-levels/}, we set the high-noise background level at 83 dB(A), which is common in busy public places such as noisy restaurants \cite{to2014noise}. The low-noise background was adjusted at 60 dB(A), which corresponds to a normal conversation sound level.


\subsection{Loudness Matching Task}
\begin{figure}[t!] 	
\centering
\includegraphics[width=0.82\linewidth]{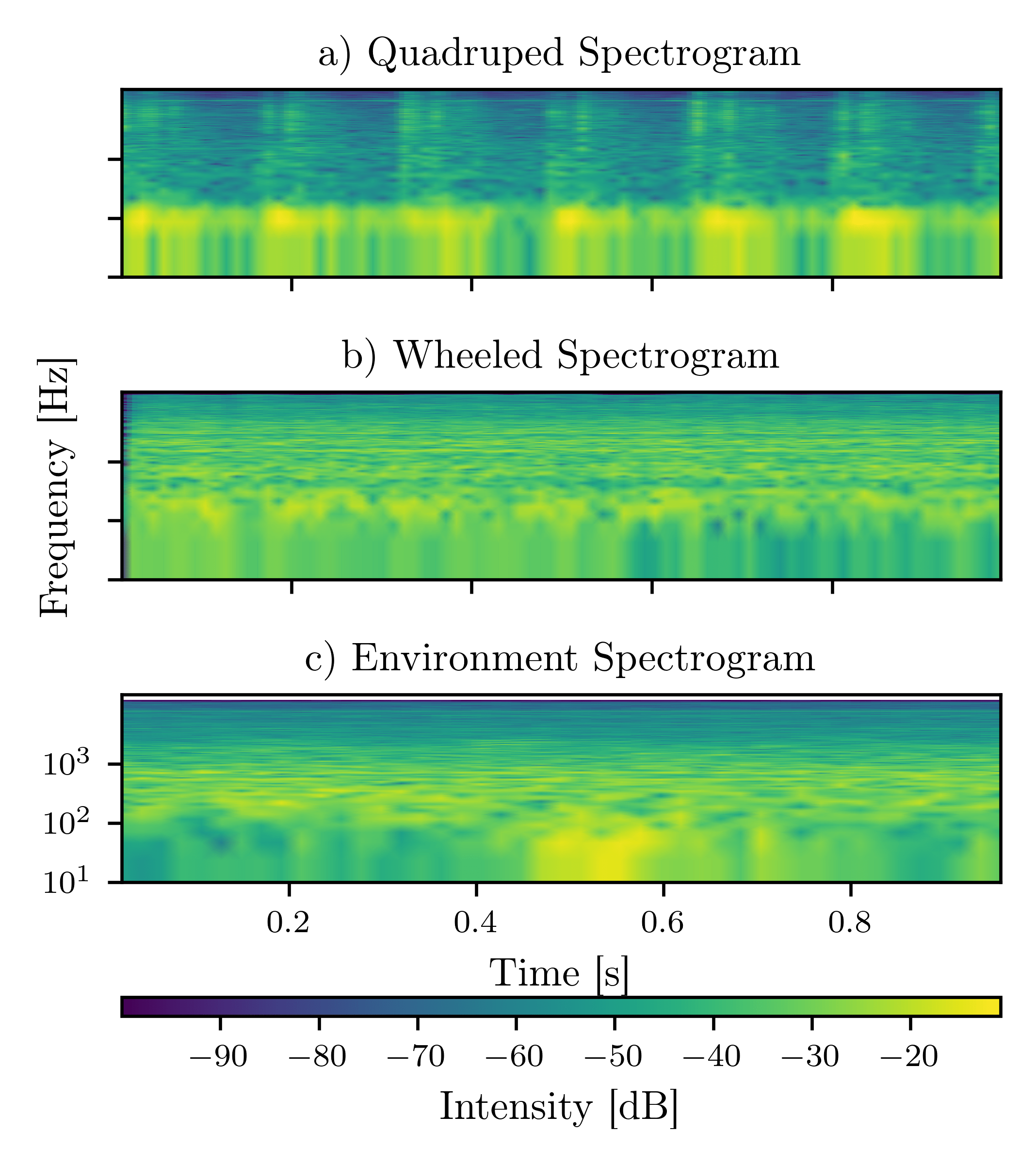}
\caption{Spectrogram of the sound profiles of a) the quadruped robot, showing interleaved high energy moments observed in the bands shown,
b) the wheeled robot, which is similar in terms of energy spread over time as environment sound depicted in c).}
\label{fig:sound_profiles}
\end{figure}

%
Our study sought to determine the distance at which humans reliably detect robots emitting two distinct consequential sounds. 
These sounds clearly differed with respect to their sound profile (e.g., temporal sound structure, Fig. \ref{fig:sound_profiles}) but may also be differently perceived in terms of loudness. 
Therefore, we ensured the same perceived loudness of the robot sounds in the detection task, allowing to unambiguously link potential detectability differences between the robots to their sound profile and rule out that a potential effect is solely driven by loudness differences \cite{oberfeld_overestimated_2022}. 
To do so, it was not sufficient to reproduce the robot sound at identical physical intensities because the subjectively perceived loudness could still differ \cite{hartmann_signals_1998}. 
Therefore, we conducted a loudness matching task prior to the detection task, in which each participant matched the loudness of the quadruped robot sound ('\textit{comparison}') to that of the wheeled robot sound ('\textit{standard}'). \\
Participants listened to pairs of robot sounds through headphones and indicated via button press which sound they perceived as louder. 
A pair consisted of the sound of the wheeled and the quadruped robot.
Each trial presented both sounds for 1500 ms with a silent inter-stimulus interval of 1000 ms. 
%
Using an adaptive procedure, we adjusted the sound level of the \textit{comparison} in accordance with each participants’ responses. 
We maintained the original wheeled sound level for the \textit{standard}.
We employed two randomly interleaved adaptive tracks for the loudness matches, for which we varied whether the \textit{comparison} was presented first or second within a trial. 
The two adaptive tracks followed a 1-down, 1-up rule \cite{levitt_transformed_1971}: 
if the participants responded that the \textit{comparison} was louder than the \textit{standard}, the level of the \textit{comparison} presented on the subsequent trial was decreased; it was increased otherwise. 
This adaptive rule tracked the 50\%-point on the psychometric function, which reveals the stimulus intensity at which participants are equally likely to respond that the wheeled or quadruped robot is perceived as louder (point of subjective equality). 
The initial level adjustments used larger steps ($\pm$5 dB) to rapidly approximate the matching loudness level. 
After 4 initial reversals (i.e., change in direction of stimulus adjustment) with larger step size, the adjustments were reduced to $\pm$2 dB for precise estimation. 
The track ended either when another 12 reversals had occurred or 50 trials had been presented. 
For each adaptive track, the initial sound levels of the robot sounds were those originally recorded.\\
%
For each adaptive track and each participant, we calculated the arithmetic mean of the \textit{comparison} (quadruped robot) levels at the maximum possible even number of reversals with small step size to determine the required change in sound level (gain) for matching the loudness of the \textit{standard} (wheeled robot). 
If the number of reversals with small step size was odd, we excluded the first reversals from the analysis. 
All tracks included $\geq$ 4 reversals with small step size or a standard deviation of \textit{comparison} levels at the reversals with small step size $<$ 5 dB, which was a-priori defined as exclusion criterion.
We averaged the determined individual gains of the two tracks per participant, if available. 
The established individual gains were used to set the sound levels for the quadruped robot in the detection task.

\subsection{Auditory Robot Detection Task}

The detection task determined the distance at which listeners reliably detected a robot sound in background noise (“detection distance”) in 8 experimental conditions, which resulted from the combinations of the factors \textit{robot sound} (wheeled robot at original level, quadruped robot at individually loudness matched level), \textit{background noise} (low = 60 dB, high = 83 dB), and \textit{secondary task} (absent, present). 
In each trial, participants were presented with two stimulus intervals (1500 ms each, with a 1000 ms silent interval between) via  headphones (Fig.~\ref{fig:user_study_design}). 
Both stimulus intervals contained background noise, while we randomly selected which one also contained the robot sound.
Participants were tasked to indicate via button press which of the two intervals contained the robot sound. 
The detection task was divided into two blocks, one with and one without secondary task. 
The order of these blocks was counterbalanced, i.e., half of the participants started with the block with secondary task, while the other half started with the block without it. 
Within each block, the combinations of robot sound and background noise were presented in random order. 
All participants completed all experimental conditions (within-subjects design). \\
For each participant and each experimental condition, we applied a 3-down, 1-up adaptive rule to determine the sound intensity of the robot at which listeners correctly detected the robot with a probability of 79.4\% \cite{levitt_transformed_1971}. 
This also means that in 20.6\% of the cases, a listener would not detect robot. 
If a participant detected the robot correctly in three consecutive trials of one adaptive track (combination of experimental condition and adaptive rule), the sound intensity was decreased in the subsequent trial. 
However, each incorrect response resulted in an increase in sound intensity in the next trial (Fig.~\ref{fig:user_study_design}b).
Note that we adjusted the sound intensity in each adaptive track by applying a linear function of distance.
In a real free field, the sound intensity relates, however, to a logarithmic function of distance. 
We therefore log-transformed the sound intensity after the adaptive procedure to consider this relationship adequately.
Nonetheless, the adaptive parameters were based on the sound intensity, which corresponded to the non-transformed distances, referred to as \textit{$D_\textit{NT}$}.
\textit{$D_\textit{NT}$} in each adaptive track was initially set to 50 and was subsequently adjusted with two relative step sizes. 
Until 4 reversals had occurred, \textit{$D_\textit{NT}$} of the previous trial was multiplied with or divided by $f = 0.7$ (for increasing or decreasing the sound intensity in the next trial, respectively). 
The step size was then reduced ($f = 0.9$) to estimate the respective 79.4\%-point more precisely.
A track ended after 8 reversals with smaller steps size or after 60 trials. 
For each adaptive track and each participant, we calculated the arithmetic mean of \textit{$D_\textit{NT}$} at the maximum possible even number of reversals with small step size. 
Applying a transformation to \textit{$D_\textit{NT}$} according to the inverse square law of sound intensity \cite{hartmann_signals_1998}, we determined the detection distance.
Again, this revealed the distance at which listeners reliably detected a robot sound in background noise. 
If the number of reversals with small step size was odd, we excluded the first of those reversals from the analysis. 
Tracks with $<$ 4 reversals with small step size were excluded from analysis. 
In total, $n = 8$ tracks were excluded (5.56\% of 144 tracks in total; total number of recorded trials $n = 9360$). 
Note that we slightly adjusted the procedure after 9 participants: We 
1) reduced the required number of reversals with larger step size from 4 to 3, 
2) decreased the initial distance of the robot from 50 to 35 m, and 
3) increased the maximum trial number per track 60 to 70. 
These changes enhanced the chance to achieve a higher number of reversals with the smaller step size, thus promoting the precision of the point estimator in each experimental condition, but did not introduce systematic differences between the experimental conditions.\\
The derived detection distances were analyzed with a linear mixed-effects model \cite{pinheiro_nlme_2000} (LMM, created with \textit{lme()}-function, R package \textit{nlme} \cite{pinheiro_nlme_2021}) with \textit{robot sound}, \textit{background sound} and \textit{secondary task} as categorical fixed effects and \textit{participant} as random intercept to control for repeated measurements. 
We used the restricted maximum-likelihood method to estimate the variance components and the Kenward-Roger approximation of the degrees of freedom.

\subsection{Secondary Task}
The purpose of the secondary task was to manipulate the participants’ cognitive workload, allowing us to examine its impact on the detectability of robot sounds in the primary task. This made our findings more applicable to real-world situations, where people may not be exclusively focused on the robot but could be engaged in other activities.
During one of the two detection task blocks (randomly assigned), all participants additionally completed an n-back task ($n$ = 1, \cite{conway_working_2005, kirchner_age_1958}) which requires to continuously indicate whether each presented letter in a continuous stream of single letters matched the immediately preceding one via button press (stimulus duration = 1.7 s, inter-stimulus interval = 0.3 s). 

\subsection{Participants}
19 individuals participated in the study in exchange for a monetary reward (15 €/h). All reported having (corrected-to-)normal hearing and vision. During testing of one participant, technical issues occurred, so that data of 18 participants (4 women, 14 men; $M_{age}$ = 25.88 years, $SD_{age}$ = 3.07 years) were available for analyses. 
The study was in line with the principles of the Helsinki Declaration and was approved by the Ethics Committee of Rheinische Friedrich-Wilhelms-Universität Bonn (approval number: 048/20). 

\subsection{User Study Procedure}
We conducted a sound test to ensure the equipment functionality in each session.
Prior to testing, all participants received study information, gave written consent, and answered demographic questions. 
Before the participants completed the loudness matching task, they had the opportunity to familiarize themselves with the robot sounds and the task in a short training. 
Subsequently, participants took a short break while the experimenter derived the individual loudness gain, which was utilized in the following detection task.\\
The detection task, the second and main part of the study, was divided into two experimental blocks (one with and one without secondary task, as described earlier). 
Again, participants completed a short training to familiarize themselves with the task and subsequently completed the experimental trials. 
After each block, participants answered the questions of the NASA Task Load Index questionnaire (NASA TLX) \cite{hart1988development} to measure their cognitive load during the two blocks, mainly serving as manipulation check. 
Finally, we measured how annoying the robot sounds were perceived by the participants, using a 7-point rating scale ranging from 1 = “not annoying at all” to 7 = “very annoying”. 
The ratings indicated higher values for the wheeled (\textit{M} = 4.39, \textit{SD}=1.82) than for the quadruped robot (\textit{M} = 2.61, \textit{SD} = 1.58).
On average, the experimental procedure took approximately 1.25 h per participant.

\begin{table*}[!t]
\centering
\begin{tabular}{rcccccc}
\toprule
Predictor & $\beta$ & $SE$ & $df$ & $t$ & $p$ & $R^2_\textit{partial}$\\
\midrule
(Intercept) & $92.60$ & $2.30$ & $110$  & $40.30$ & $<.001$ & $ $\\
Background Noise & $-27.72$ & $1.52$ & $110$ & $-18.26$ & $\mathbf{<.001}$ & $0.67$\\
Robot Sound & $-26.23$ & $1.52$ & $110$ & $-17.28$ & $\mathbf{<.001}$ & $0.64$\\
Cognitive Task & $0.49$ & $1.52$ & $110$ & $0.32$ & $.749$ & $0$\\
Background Noise $\times$ Robot Sound & $22.79$ & $2.24$ & $110$ & $10.19$ & $\mathbf{<.001}$ & $0.38$\\
Background Noise $\times$ Cognitive Task & $-0.72$ & $2.13$ & $110$ & $-0.34$ & $.737$ & $0$\\
Robot Sound $\times$ Cognitive Task & $-0.83$ & $2.13$ & $110$ & $-0.39$ & $.698$ & $0$\\
Background Noise $\times$ Robot Sound $\times$ Cognitive Task & $1.41$ & $3.12$ & $110$ & $0.45$ & $.652$ & $0$\\
\bottomrule
\end{tabular}
\caption{
        Estimated fixed effects parameters of the LMM predicting the individual detection distances from the background noise, robot sound and cognitive task. Shown are effect estimates ($\beta$), standard errors ($SE$), degrees of freedom ($df$), $t$-values, $p$-values, semi-partial $R^2$ for each fixed effect ($R^2_\textit{partial}$).
        \label{tab:LMMresults}
        }
\end{table*}
\section{Results and Discussions}
\label{sec:results}

All statistical tests reported in the following were two-sided and considered $p$-values of .05 as cut-off. 

\subsection{Loudness Matching Results}
\begin{figure}[h!]
\centering
\includegraphics[width=0.7\linewidth]{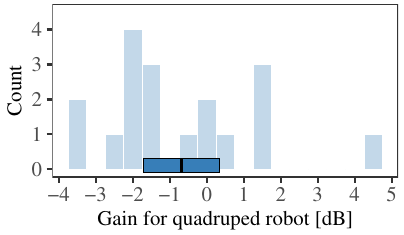} 	
\caption{ 		
        Histograms of the individual gains in dB (light blue bars) for the quadruped robot to match the loudness of the wheeled robot at a distance of 1 m (bin width = 0.5 dB). \\The dark blue horizontal box indicates the 95\% confidence interval, the vertical line within the confidence interval represents the mean. Note that the mean was not significantly different from 0, indicating that both sounds were perceived to be rather similarly loud.
		\label{fig:loudnessmatch}
	} 
 \end{figure} 
The gains (sound level change) required for the quadruped robot's sound to match the perceived loudness of the wheeled robot are shown in Fig.~\ref{fig:loudnessmatch}.
The majority of the participants (66.67\%) perceived the quadruped as (slightly) louder than the wheeled robot (negative gains indicate that the sound level was required to be reduced for the loudness match). 
A paired samples \textit{t}-test revealed, however, that the determined gains were, on average, not significantly different from 0, $\textit{t}(17) = 1.41, \textit{p} = .176$, suggesting that the \textbf{loudness of the two robot sounds was perceived comparably}.


\subsection{Manipulation Check}
In principle, it was possible that participants did not experience the secondary task as cognitively demanding or neglected the secondary task and only focused on the detection task, which could potentially explain a non-significant impact of the secondary task on \textit{$D_\textit{detect}$}. 
This explanation could, however, be ruled out. 
The NASA-TLX determined the magnitude of cognitive load in the block with and without secondary task (score between 0 and 100). It indicated that the cognitive load was significantly higher in the block with ($M = 67.24$, $SD = 12.84$) than without secondary task ($M = 55.98$, $SD = 14.40$),  \textit{t}-test ($\textit{t}(17) = 3.65, \textit{p} = .002$).
At the same time, participants achieved $88.7\%$ - $99.7\%$ correct responses ($M = 96.5\%$), underscoring high performance levels. 
In sum, the performance in the secondary task was high and the \textbf{secondary task increased the cognitive load significantly}, as supposed. \\
However, it is worth noting that in actual HRI, individuals may find themselves immersed in even more cognitively challenging tasks. 
Therefore, it cannot be dismissed that higher cognitive load levels than introduced by the secondary task in this study could indeed influence the detectability of robot sounds, which might be addressed in future studies.


\subsection{\textcolor{black}{Detection Distances}}
\begin{figure}[h!] 	
\centering 	
\includegraphics[width=0.8\linewidth]{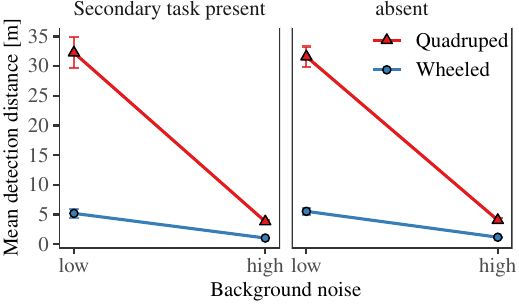} 	
\caption{ 		
        Mean detection distances for the two robot sounds as a function of background noise (\textit{x}-axis) and secondary task (left: present, right: absent). Red triangle: Quadruped robot. Blue circle: Wheeled robot. Error bars indicate $\pm1$ SE of the individual means.
		\label{fig:detectionDistance}
	} 
 \end{figure} 
The detection task revealed the sound intensity with which a robot could be reliably detected by its sound. 
According to the inverse square law, we calculated the distances corresponding to the determined sound intensities, the detection distances \textit{$D_\textit{detect}$}. 
We analyzed potential effects of the robot sound, background sound and secondary task on \textit{$D_\textit{detect}$}, using a LMM (Table~\ref{tab:LMMresults}). 
The model explained 89.10\% of the variance with its fixed and random effects.\\
The participants detected the quadruped robot at a significantly larger distance than the wheeled robot (Fig.~\ref{fig:detectionDistance}).
On average, the wheeled robot had to approach the participant as close as 3.45 m (\textit{SD} = 1.24 m) to be reliably detected. 
In contrast, a distance of 17.80 m (\textit{SD} = 4.09 m) was sufficient for the participant to reliably detect the quadruped robot. 
In addition, the robots had to be closer to the participant to be reliably detected in high (\textit{M} = 2.75 m, \textit{SD} = 0.71 m) than in low background noise (\textit{M} = 18.60 m, \textit{SD} = 4.68 m).
These results suggested that humans \textbf{detected the quadruped robot more easily} (i.e., at larger distances) than the wheeled robot, and that they \textbf{detected a robot in a quieter environment more easily} than in a noisy one, which were \textbf{both mostly independent of their engagement in a secondary task}.\\
The effect of the background noise was substantially more pronounced for the quadruped than for the wheeled robot (\( \text{robot sound } \times \text{ background noise}\)), indicating that the detectability for the quadruped suffered more strongly from the increase in background noise. 
Nonetheless, two-sided contrasts with model-based standard error (SE) and Bonferroni-correction confirmed that the difference between the robot sounds was significant in both background noise conditions (both \textit{p}$ <.012$). 
In sum, the \textbf{wheeled robot had to be closer than the quadruped robot to be reliably detectable, even in high background noise.}
Finally, the differences between \textit{$D_\textit{detect}$} for the two robot sounds and the two background noise levels were quite similar in conditions with and without the secondary task, as can be seen in Fig.~\ref{fig:detectionDistance}. The non-significant interactions \( \text{robot sound} \times \text{secondary task} \) and \( \text{background noise} \times \text{secondary task} \) confirmed that the two reported main effects of robot sound and background noise were not substantially dependent of the engagement in a secondary task. 

Our findings allow for the following implications for practical applications and future research: 
Humans are more sensitive to the sound of a quadruped than to that of a wheeled robot, which may affect other aspects of HRI as well (e.g., the distance humans would like to maintain to a robot).
This large relative difference in sensitivity underscores the need of considering the robot type and their consequential sound when developing social navigation algorithms for mobile robots. 
It also suggests that in situations where people prefer the robot to remain rather unnoticed, the wheeled robot seems to be a suitable candidate due to its low detection distance.
In contrast, when humans collaborate with robots and prefer to detect the robot reliably, even when out of sight, the use of the quadruped robot would be better suited.
However, it remains to be investigated which specific acoustic cues improve the detectability. This would allow to make presumptions about other existing robots merely based on a sound analysis and to inform future robot sound design.
The background noise is an additional factor that should be considered in developing social navigation algorithms. 
Nonetheless, the sensitivity difference between the robots persisted in high background noise, such that the outlined implications are also valid for high-noise environments, such as industrial settings.
The difficulty of developing HRI in noisy environments described by Tsarouchi \etal~\cite{tsarouchi2016human} could therefore potentially be solved by using the quadruped robot in high background noise.
However, it remains to be investigated if the consequential sound of a quadruped robot in a noisy environment is sufficient to enable precise auditory localization of the robot or whether it impacts the distance humans prefer to keep from a robot \cite{leichtmann2020much}. 
These aspects also seem crucial to be taken into account when designing human-centered HRI concepts because they contribute to a safe and from a user perspective pleasant and enjoyable human-robot interplay in different contexts.
Furthermore, future research could address the potential and requirements of dynamic navigation algorithms, which consider the robot's speed, path and operational mode.

\section{\textcolor{black}{Conclusion}}
\label{sec:conclusion}

In summary, our study investigated the auditory detectability of two distinct types of mobile robots operating beyond human sight: a wheeled (Turtlebot 2i) and a quadruped robot (Unitree Go 1).
In a controlled experiment, we determined for the first time at which distances humans can reliably detect the robots sounds with varying background noise levels and levels of cognitive engagement.
Our results showed a lower auditory detectability in noisier environments, but most importantly, they demonstrated that the wheeled robot had to be significantly closer the quadruped robot for reliable detection, even in noisy environments.
Overall, our findings provide implications for the improvement of social robot navigation and the definition of an optimal scope of use contexts for each of the investigated robots, contributing to the human-centered and context appropriate design of HRI.

\section*{Acknowledgments}
We thank Prof. Dr. Caja Thimm for letting us borrow the equipment and recording studio at the Medienwissenschaften, University of Bonn, Germany.

\bibliographystyle{IEEEtran}
\bibliography{main-bibliography,jorge_paper,MarleneBib,software}

\end{document}